\documentclass{article} 
\usepackage{nips14submit_e,times}
\usepackage{authordate1-4}
\usepackage{hyperref}
\usepackage{url}

\title{Substitute Based SCODE Word Embeddings \\ in Supervised NLP Tasks}

\author{Volkan Cirik \quad Deniz Yuret\\
Artificial Intelligence Laboratory \\
Ko\c{c} University, \.Istanbul, Turkey \\
\texttt{\{vcirik,dyuret\}@ku.edu.tr}
}

\nipsfinalcopy 

\begin{document}

\maketitle

\begin{abstract}
We analyze a word embedding method in supervised tasks.  It maps words on a sphere such that words co-occurring in similar contexts lie closely. The similarity
of contexts is measured by the distribution of substitutes that can fill them. We compared word embeddings, including more recent representations \cite{huang2012,mikolov2013}, in Named Entity Recognition (NER), Chunking, and Dependency Parsing. We examine our
framework in multilingual dependency parsing as well. The results show that the proposed method achieves as good as or better results compared to the other word embeddings in the tasks we investigate. It achieves state-of-the-art results in multilingual dependency parsing. Word embeddings in 7 languages are available for public use\footnote{https://github.com/ai-ku/wvec/}.
\end{abstract}

\section{Introduction}
\label{sec:intro}
Word embeddings represent each word with a dense, real valued vector.
The dimension of word embeddings are generally small compared
to the vocabulary size. They do not suffer from sparsity unlike
one-hot representations which have the dimensionality of
the vocabulary and a single non-zero entry. 
They capture semantic and syntactic similarities \cite{mikolov2013}.
They may help reduce the dependence on hand-designed features 
which are task and language dependent. 


We analyze a word embedding method proposed in \cite{yatbaz2012}, in supervised Natural Language Processing (NLP) tasks. The method represents the context of a word by its probable substitutes. Words with their probable substitutes are fed to a co-occurrence modeling framework (SCODE) \cite{maron2010}. 
Words co-occurring in similar context are closely embedded on a sphere. These word embeddings achieve state-of-the-art results in inducing part-of-speech (POS) tags for several languages \cite{yatbaz2014}. However, their use in supervised tasks has not been well studied so far. This study aims to fill this gap.

\cite{turian2010} compared word embeddings in Named Entity Recognition (NER) and Chunking. They use word embeddings as auxiliary features in existing systems. They improved results in both tasks compared to the baseline systems. Following this study, we report results in Chunking and NER benchmarks for SCODE embeddings. In addition, we examine word embeddings in dependency parsing. We report multilingual dependency parsing results for SCODE embeddings as well.

SCODE embeddings achieve comparable or better results compared to the other word embeddings. Multilingual results in dependency parsing also suggest that SCODE embeddings are consistent in achieving good results across different languages.

\section{Related Work}

In this section, we introduce word embeddings we mentioned in this work.

\begin{itemize}
\item {\bf C\&W:} \cite{collobert2008} introduce a convolutional neural
network architecture that is capable of learning a language
model and generating word embeddings from unlabeled data. The model
can be fine-tuned for supervised NLP tasks. 

\item {\bf HLBL:} \cite{mnih2007} introduce the log-bilinear language model. It is a feed-forward neural network with one linear hidden
layer and a softmax output layer.  The model utilizes
linear combination of word type representations of preceding
words to predict the next word.  \cite{mnih2009} modify this model
to reduce computational cost by introducing a hierarchical structure. The
architecture is then named the hierarchical log-bilinear language model.

\item {\bf GCA NLM:} \cite{huang2012} introduce an architecture using both local and global context via a joint training objective. The training is very similar to \cite{collobert2008}. They represent
a word context  by taking the weighted average of the representations of word types in a fixed size window around the target word token. Following \cite{reisinger2010},
they cluster  word context representations for each word type
to form word prototypes. These prototypes capture homonymy and polysemy relations.

\item {\bf LR-MVL:} \cite{dhillon2011} present a spectral method to induce word embeddings. They perform the Canonical Correlation Analysis on the context
of a token. They provide an algorithm to represent a target word with different vectors depending on its context. The objective function they define is convex. Thus, the method is guaranteed to converge to the optimal solution. 

\item {\bf Skip-Gram NLM:} \cite{mikolov2010} propose a two neural models to induce word embeddings. The first architecture is Continuous Bag-of-Words where the words in a window surrounding the target is used to classify the target word. The second one is continuous Skip-Gram model in which the target word is used to classify its surrounding words. \cite{mikolov2013} show that these representations reflect syntactic and semantic regularities.

\item {\bf SCODE Word Embeddings:} \cite{maron2010} introduce the SCODE framework, an extension of the
CODE \cite{globerson2007} framework. \cite{maron2010} obtains word type
representations from co-occurrence data generated by using neighbors of words.  \cite{yatbaz2012} extend this work by generating co-occurrence data using probable substitutes of words. In Section ~\ref{sec:scode-embeddings}, we explain this framework in detail. Here, we review studies extending that work.

\cite{baskaya2013ai} used SCODE word embeddings for Word Sense
Induction. They achieved the best results in Semeval 2013 Shared Task \cite{jurgens2013semeval}. \cite{cirik2013ai} is the first study exploiting SCODE embeddings in a supervised setup by using them as word features. 

\end{itemize}

\section{Substitute Based SCODE Word Embeddings} \label{sec:scode-embeddings}

In this section, we summarize our framework based on \cite{yatbaz2012}. In Section ~\ref{sec:substitute}, we explain substitute word distributions. In Section ~\ref{sec:discretization}, we explain how substitute word distributions are discretized. In Section ~\ref{sec:scode} we introduce Spherical Co-Occurrence Data Embedding framework \cite{maron2010}. 

\subsection{Substitute Word Distributions} \label{sec:substitute}

Substitute word distributions are defined as the probability of observing
a word in the context of the target word. We define the context of a target word as the sequence of words in the window of size $2n - 1$ centered at the position of the target word token. The target word is excluded in the context. 

\begin{quote}
(1)``Steve Martin has already {\bf laid} his claim to that ." 
\label{sentenceExample}
\end{quote}

For example, in the sentence (1),  the context of the word token `\emph{laid}', for $n =4$, is `\emph{ Martin has already  --- his claim to }' where \emph{---} specifies the position of the target word token.

\begin{table}[h]
\small
\caption{\label{vector-table} Substitute word distribution for ``laid'' in  sentence (1). }
\begin{center}
\begin{tabular}{lrlll}
\textbf{Probability} & \textbf{Substitute Word}&  &  &  \\ \cline{1-2}
0.191       & staked         &  &  &  \\
0.161       & established            &  &  &  \\
0.125       & made            &  &  &  \\
0.096       & proved            &  &  &  \\
0.094       & rejected        &  &  &  \\
\end{tabular}
\end{center}
\end{table}

 Table~\ref{vector-table} illustrates the substitute 
 distribution of ``laid'' in (1). There is a row for each word in the vocabulary. For instance, probability of ``established'' occurring in the position of ``laid'' is 0.161 in this context.
 
 Let target word token be in the position $0$, the context
spans from positions $-n + 1$ to $n - 1$. The probability of
observing each word  $w$ in vocabulary in the context of the target
word token is calculated as follows:

\begin{eqnarray}
  \label{eq:lm1}P(w_0 = w | c_w) & \propto & P(w_{-n+1}\ldots w_0\ldots w_{n-1}) \\
  \label{eq:lm2}& = & P(w_{-n+1})P(w_{-n+2}|w_{-n+1})\nonumber\\
  &&\ldots P(w_{n-1}|w_{-n+1}^{n-2})\\
  \label{eq:lm3}& \approx & P(w_0| w_{-n+1}^{-1})P(w_{1}|w_{-n+2}^0)\nonumber\\
  &&\ldots P(w_{n-1}|w_0^{n-2})
\end{eqnarray}

In the Equation \ref{eq:lm1}, the right-hand side is proportional
to the left-hand side because $P(c_{w_{0}})$ is independent of any
word $w$ for $w_{0}$.  After using the chain rule,
Equation \ref{eq:lm2} is obtained from the right-hand side of Equation \ref{eq:lm1}. By applying $n^{th}$-order Markov assumption, only the closest $n - 1$ words in each term of the Equation \ref{eq:lm2} are needed which equals to the Equation \ref{eq:lm3}.  The Equation \ref{eq:lm3} is proportional to the Equation \ref{eq:lm2} because the context of the target word is fixed, thus, any term that does not depend on $w_{0}$ is fixed.  Equation \ref{eq:lm3} are truncated or dropped near the boundaries of the sentence. (e.g. if $0$ is the first word of a sentence, $P(w_{0} | w_{-n+ 1}^{-1})$ becomes $P(w_{0})$). An n-gram language model provides the probabilities required
for Equation \ref{eq:lm3}. 

\subsection{Discretization of Substitute Word Distributions} \label{sec:discretization}

The co-occurrence embedding algorithm we describe in Section ~\ref{sec:scode}, requires its input as categorical variables co-occurring together. We aim to associate words co-occurring in the same context. Although substitute word distributions represent the context of a word, they are categorical probability distributions. Thus, they should be transformed into a discrete setting.

We sample word types from substitute word distributions. The number of samples should be chosen carefully, if the number of the samples are too small, it may fail to capture the characteristics of the distribution.

Figure~\ref{fig:sampling-example} is an example of a discretization with sampling. Substitute words are sampled from substitute word distributions of sentence (1).

\begin{figure}
  \centering

\begin{tabular}{lr}
\textbf{Word Token} & \textbf{Substitute Word} \\ \hline
Steve   & Mr        \\
Steve   & Chris     \\
Martin  & Coppell   \\
Martin  & Wilson    \\
has     & had       \\
has     & has       \\
already & finally   \\
already & already   \\
laid    & made      \\
laid    & shown     \\
his     & no        \\
his     & no        \\
claim   & response  \\
claim   & testimony \\
to      & to        \\
to      & to        \\
that    & fame      \\
that    & succeed   \\
.       & .         \\
.       & .                             
\end{tabular}
  \caption[Co-occurrence data from sampling discretization]{Sampling
    twice from the substitute word distributions of sentence (1).}
  \label{fig:sampling-example}
\end{figure}

\subsection{Spherical Co-Occurrence Data Embedding} \label{sec:scode}

This section shortly reviews the Symmetric Interaction Model of the Co-occurrence Data Embedding (CODE) \cite{globerson2007} and its extension Spherical Co-Occurrence Data Embedding (SCODE) \cite{maron2010}. 

We map co-occurrence data generated from the word types and substitute word distributions described in Section \ref{sec:discretization} to $d$ dimensional Euclidean space.  

Let $X$ and $Y$ have a joint distribution such that $X$ and $Y$ are categorical variables with finite cardinality
$|X|$ and $|Y|$.  However we only observe a set of pairs $\{x_i, y_i\}_{i=1}^n$
drawn IID from the joint distribution of $X$ and $Y$.  These pairs are
summarized by the empirical distributions $\bar{p}(x,y)$, $\bar{p}(x)$
and $\bar{p}(y)$.  Embeddings
 $\phi(x)$ and $\psi(y)$ can capture the statistical relationship
between the variables $x$ and $y$ in terms of square of Euclidean distance \mbox{$d^2_{x,y} =  \|\phi(x) - \psi(y)\|^2$}. In other words, pairs
frequently co-occurring are embedded closely in $d$ dimensional space.

We used the following extended model \cite{maron2010} proposed among others in \cite{globerson2007} :
\begin{equation} \label{eq:marginal-marginal-model}
  p(x,y) = \frac{1}{Z} \bar{p}(x) \bar{p}(y) e^{-d^2_{x,y}}
\end{equation}
where $Z = \sum_{x,y} \bar{p}(x) \bar{p}(y) e^{-d^2_{x,y}}$ is the
normalization term.  The log-likelihood of the joint
distribution over all embeddings $\phi$ and $\psi$ can be described as the following:

\small
\begin{eqnarray} \label{eq:log-likelihood}
&& \ell(\phi,\psi) =  \sum_{x,y} \bar{p}(x,y) \log p(x,y) \\
&& =  \sum_{x,y} \bar{p}(x,y) (-\log Z + \log \bar{p}(x) \bar{p}(y) - d^2_{x,y})  \\
&& =  -\log Z + const - \sum_{x,y} \bar{p}(x,y) d^2_{x,y}
\end{eqnarray}
\normalsize

The gradient of the log-likelihood depends on the sum of embeddings
$\phi(x)$ and $\psi(y)$, for $x \in X$ and $y \in Y$, and to maximize
the log-likelihood, \cite{maron2010} use a gradient-ascent
approach. The gradient is :
\footnotesize
\begin{equation} \label{eq:gradient-phi}
\frac{\partial \ell(\phi,\psi)}{\partial \phi(x)} =
\sum_{y} 2 \bar{p}(x,y) {[}\psi(y) - \phi(x){]} \\
+ \frac{1}{Z} \sum_{y} \bar{p}(x) \bar{p}(y) {[}\phi(x) - \psi(y){]} e^{-d^2_{x,y}}
\end{equation}
\begin{equation} \label{eq:gradient-psi}
  \frac{\partial \ell(\phi,\psi)}{\partial \psi(y)} =%
  \sum_{x} 2 \bar{p}(x,y) {[}\phi(x) - \psi(y){]} \\ +%
  \frac{1}{Z} \sum_{x} \bar{p}(x) \bar{p}(y) {[}\psi(y) - \phi(x){]} e^{-d^2_{x,y}}
\end{equation}
\normalsize 
The first sum in (\ref{eq:gradient-phi}) and (\ref{eq:gradient-psi}), the gradient of the part with $d^2_{x,y}$ of (\ref{eq:log-likelihood})  acts as an attraction force between the $\phi(x)$  and $\psi(y)$.  The second sum in (\ref{eq:gradient-phi}) and (\ref{eq:gradient-psi}) , the gradient of $-\log Z$ in
(\ref{eq:log-likelihood}) acts a repulsion force between the
$\phi(x)$ and $\psi(y)$. 

\cite{maron2010} constrain all embeddings $\phi$ and
$\psi$ to lie on the $d$ dimensional unit sphere, hence the name
SCODE.  A coarse approximation in
which all $\phi$ and $\psi$ distributed uniformly and independently on
the sphere, enables $Z$ to be approximated by a constant value.
Thus, it does not require the computation of $Z$ during training.

For the experiments in the work, we use SCODE with sampling based
stochastic gradient ascent a constant approximation of $Z$ and
randomly initialized $\phi$ and $\psi$ vectors. 

\section{Induction of Word Embeddings}  \label{sec:inducing}
This section explains how we induced Substitute Based SCODE Word Embeddings and obtain other embeddings.  We report the details of unlabeled data used to induce word embeddings. We present the parameters chosen for induction. We explain how we obtain other word embeddings.

\subsection*{Unlabeled Data}  \label{sec:scode-unlabeleddata}

Word embeddings require large amount of unlabeled data to efficiently capture syntactic and semantic regularities. The source of the data also may have an impact on the success of the word embedding on the labeled data. Thus, we induce word embeddings using a large unlabeled corpora.
 
Following \cite{turian2010}, we used RCV1 corpus containing 190M word tokens \cite{rose2002reuters} corpus. We removed all sentences that are less than 90\% lowercase a–z. The whitespace is not counted. After following the preprocessing technique described in \cite{turian2010}, the corpus has 80M word tokens.


We induce word embeddings for multilingual experiments explained in Section ~\ref{sec:depparsing1}. We generate embeddings using subsamples of corresponding Tenten Corpora \cite{jakubivcek2013tenten} for Czech, German, Spanish and Swedish and Wikipedia dump files for Bulgarian, Hungarian. For Turkish, we used a web corpus \cite{sak2008turkish}. Table ~\ref{corpus-table} shows the statistics of unlabeled corpora for languages.
\begin{table}[h]
\small
\caption{\label{corpus-table} Unlabeled Corpora of Different Languages for Word Embeddings }
\begin{center}
\begin{tabular}{llr}
\textbf{Language}  & \textbf{Corpus}     & \textbf{Number Of Words} \\ \hline
Bulgarian & Wikipedia  & 101M            \\
Czech     & Tenten     & 140M            \\
English   & RCV1       & 80M             \\
German    & Tenten     & 180M            \\
Spanish   & Tenten     & 106M            \\
Swedish   & Tenten     & 113M            \\
Turkish   & Web Corpus & 180M           
\end{tabular}
\end{center}
\end{table}
\begin{table*}[!b!t!h]
\small
\caption{\label{table:coverage} Word token coverage for word embeddings. }
\begin{center}
\begin{tabular}{lrrrrrrrr}
 & \multicolumn{3}{l}{Chunking} & \multicolumn{3}{l}{NER} & \multicolumn{2}{l}{Dependency Parsing} \\
\multicolumn{1}{l|}{Word Embeddings} & \multicolumn{1}{l}{Training} & \multicolumn{1}{l}{Development} & \multicolumn{1}{l|}{Test} & \multicolumn{1}{l}{Training \& Development} & \multicolumn{1}{l}{Test} & \multicolumn{1}{l|}{OOD} & \multicolumn{1}{l}{Training} & \multicolumn{1}{l}{Test} \\ \hline
\multicolumn{1}{l|}{C\&W} & 0.9800 & 0.9832 & 0.9764 & 0.9402 & 0.9359 & 0.9631 & 0.9835 & 0.9856 \\
\multicolumn{1}{l|}{HLBL} & 0.9654 & 0.9675 & 0.9621 & 0.9549 & 0.9503 & 0.9777 & 0.9691 & 0.9674 \\
\multicolumn{1}{l|}{GCA NLM} & 0.8230 & 0.8271 & 0.8139 & 0.6971 & 0.6760 & 0.8208 & 0.8322 & 0.8270 \\
\multicolumn{1}{l|}{LR-MVL} & 0.9806 & 0.9839 & 0.9778 & 0.9422 & 0.9380 & 0.9637 & 0.9841 & 0.9862 \\
\multicolumn{1}{l|}{Skip-Gram NLM} & 0.9848 & 0.9877 & 0.9827 & 0.9117 & 0.9075 & 0.9614 & 0.9833 & 0.9852 \\
\multicolumn{1}{l|}{SCODE} & 0.9848 & 0.9877 & 0.9827 & 0.9117 & 0.9075 & 0.9614 & 0.9833 & 0.9852 \\

\end{tabular}
\end{center}
\end{table*}


\subsection*{Parameters}

To generate substitute word distributions, we trained a 4-gram statistical language model (LM) using SRILM \cite{stolcke2002}. We used interpolated Kneser-Ney discounting. We replaced words observed less than 2 times with an unknown tag. Table ~\ref{lm-table} shows the statistics of language model corpora\footnote{We should note that LM corpora differ from the word embedding corpora. The first one
is used to learn an LM which is then used for generating substitute words on the word
embedding corpora.} for each language. We used FASTSUBS algorithm \cite{yuret2012} to generate top 100 substitutes words and their substitute probabilities. 

We keep each word with its original capitalization. We sampled 100 substitutes per instance.  The SCODE normalization constant was set to 0.166. For multilingual experiments we used 25 dimension word embeddings. We observe no significant improvements in scores when we change the number of dimensions for SCODE embeddings.

\begin{table}[h]
\small
\caption{\label{lm-table} Unlabeled Corpora for Language Modeling}
\begin{center}
\begin{tabular}{llr}
\textbf{Language} & \textbf{Corpus} & \textbf{Number Of Words} \\ \hline
Bulgarian         & Wikipedia       & 850M                     \\
Czech             & Tenten          & 1.79B                    \\
English           & ukWac           & 2B                       \\
German            & Tenten          & 1.8B                     \\
Spanish           & Tenten          & 2.4B                     \\
Swedish           & Tenten          & 113M                     \\
Turkish           & Web Corpus      & 1.8B                    
\end{tabular}
\end{center}
\end{table}

\subsubsection*{Other Word Embeddings}

We downloaded word embeddings from corresponding studies\footnote{http://metaoptimize.com/projects/wordreprs/}\footnote{http://www.cis.upenn.edu/~ungar/eigenwords/}\footnote{http://goo.gl/ZXv0Ot}\cite{turian2010,dhillon2011,huang2012}. We should note that we do not use the context-aware word embeddings of \cite{dhillon2011}. These word embeddings are scaled with parameter $\sigma=0.1$, since \cite{turian2010} have shown that word embeddings achieve their optima at this value. We use 50-dimension of each word embeddings in all comparisons.

To induce Skip-Gram NLM embeddings, we ran the code provided on the website\footnote{https://code.google.com/p/word2vec/} of \cite{mikolov2010,mikolov2013} on the RCV1 corpus. We used Skip-Gram model with default parameters. We changed words occurring less than 2 times with an unknown tag. The performance of Skip-Gram NLM and SCODE word embeddings do not improve with scaling, thus, we use them without scaling.

We report word token coverage for word embeddings in Table~\ref{table:coverage}. For each task, an unknown word in the training or test phase is replaced with the word embedding of unknown tag. Thus, the word embedding method with high coverage suffers less from unknown words, which in turn effects its success. Table ~\ref{table:coverage} shows the word token coverage for each task and their corresponding datasets. GCA NLM has the lowest coverage in all tasks, which may explain its level of performance.

\section{Experiments}
\label{sec:tasks1}
In this section, we detail the experiments. We introduce tasks in which we compared word embeddings, the data used, and parameter choices made. We report results
for each task. 

\subsection*{Chunking}

We used CoNLL-2000 Shared task Chunking as the first benchmark \cite{conll2000}. The data is from Penn Treebank which is a newswire text from Wall Street Journal \cite{treebank3}. The training set contains 8.9K sentences. The development set contains 1K sentences and the test set has 2K.

\begin{table}[h]
\small
\caption{\label{features-chunking} Features Used In CRF Chunker }
\begin{center}
\begin{itemize}
  \item Word features: $w_{i}$ for  $i$ in \{-2,-1,0,+1,+2\},  $w_{i} \wedge w_{i+1}$ for $i$ in \{-1,0\}
     
   \item Tag features: $w_{i}$ for  $i$ in \{-2,-1,0,+1,+2\}, $t_{i} \wedge  t_{i+1}$ for $i$ in \{-2,-1,0,+1\},.  $t_{i} \wedge t_{i+1} \wedge t_{i+2}$ for $i$ in \{-2,-1,0\}. 
  
   \item Embedding features: $e_{i}[d]$ for $i$ in  \{-2,-1,0,+1,+2\},  where $d$ ranges over the dimensions of the embedding $e_{i}$.
  
\end{itemize}
\end{center}
\end{table}

We used publicly available implementation of \cite{turian2010}. 
It is a CRF based chunker using features
described in Table ~\ref{features-chunking}. The only hyperparameters of the model was L2-regularization $\sigma$ which is optimal at 2. After successfully replicating results in that work\footnote{We report our replication of results for word embeddings which differs from \cite{dhillon2011}.}, we ran experiments for new word embeddings.

In Table ~\ref{chunking-table1}, we report F1-score of word embeddings and the score of the baseline chunker that is not using word embeddings. They all improve baseline chunker, however, improvement is marginal for all of them. The best score is achieved by SCODE embeddings trained on RCV1 corpus.  


\begin{table}[!h]
\small
\caption{\label{chunking-table1} Chunking Results for Word Embeddings. The ones in bold font are the highest scores in their columns. }
\begin{center}

\begin{tabular}{lrr}
\textbf{Word Embeddings} & \textbf{Development Score} & \textbf{Test Score} \\ \hline
Baseline                 & 0.9416                     & 0.9379              \\
C\&W                     & \textbf{0.9466}                     & 0.9410               \\
HLBL                     & 0.9463                     & 0.9400              \\
GCA NLM                  & 0.9425                     & 0.9402              \\
LR-MVL                   & 0.9458                     & 0.9416              \\
Skip-Gram NLM        & 0.9400                     & 0.9402              \\
SCODE             & 0.9430                     & \textbf{0.9429}     \\
           
\end{tabular}
\end{center}
\end{table}

\subsection*{Named Entity Recognition} \label{sec:ner1}
The second benchmark is CoNLL-2003 shared task Named Entity Recognition \cite{conll2003}. The data is extracted from RCV1 Corpus. Training, development, and test set contains 14K, 3.3K and 3.5K sentences. Annotated named entities are location, organization and miscellaneous names. \cite{conll2003} details the number of named entities and data preprocessing. In addition, \cite{turian2010} evaluated word embeddings on an out-of-domain (OOD) data containing 2.4K sentences \cite{chinchor1997muc}.

\begin{table}[h]
\small
\caption{\label{features-ner} Features Used In Regularized Averaged Perceptron. Word embeddings are used the same way as in Table~\ref{features-chunking}.}
\begin{center}
\begin{itemize}
  \item Previous two predictions $y_{i-1}$ and $y_{i-2}$
  \item Current word $x_{i}$
  \item $x_{i}$ word type information : all-capitalized, is-capitalized, all-digits, alphanumeric etc.
  \item Prefixes and suffixes of $x_{i}$, if the word contains hyphens,then
  the tokens between the hyphens
  \item Tokens in the window $c=(x_{i-2},x_{i-1},x_{i},x_{i+1},x_{i+2})$
  \item Capitalization pattern in the window $c$
  \item Conjunction of $c$ and $y_{i-1}$
    
\end{itemize}
\end{center}
\end{table}

We used publicly available implementation of \cite{turian2010}. It is a regularized averaged perceptron model using features described in Table~\ref{features-ner}.  After we replicated results of that work, we ran the same experiments for new word embeddings. It is important to note that, unlike \cite{turian2010}, we did not use any non-local features or gazetteers because we wanted to measure the performance gain of word embeddings alone.  The only hyperparameter is the number of epochs for the perceptron. The perceptron stops when there is no improvement for 10 epochs on the development set. The best epoch on development set is used for the final model. 

Table ~\ref{ner-table1} summarizes the result of NER experiments. The first three rows from \cite{turian2010}, report the baseline and the best results for C\&W and HLBL embeddings. The baseline system does not use word embeddings as features. All of the word embeddings significantly improve the baseline system. SCODE embeddings trained on RCV1 corpus achieves the best score on test set and Out of Domain Test (OOD) set. Note that RCV1 corpus is the superset of NER training and test data. Thus, C\&W, HLBL and SCODE on RCV1 embeddings are from the same data source. 


\begin{table}[h]
\small
\caption{\label{ner-table1} NER Results for Word Embeddings. The ones in bold fonts are the highest scores in their columns. }
\begin{center}
\begin{tabular}{lrrr}
\textbf{Word Embeddings} & \multicolumn{1}{l}{\textbf{Development}} & \multicolumn{1}{l}{\textbf{Test}} & \multicolumn{1}{l}{\textbf{OOD}} \\ \hline
Baseline                 & 0.9003                                   & 0.8439                            & 0.6748                           \\
C\&W 200-dim             & \textbf{0.9246}                          & 0.8796                            & 0.7551                           \\
HLBL 100-dim             & 0.9200                                   & 0.8813                            & 0.7525                           \\ \hline
C\&W                     & 0.9227                                   & 0.8793                            & 0.7574                           \\
HLBL                     & 0.9146                                   & 0.8705                            & 0.7293                           \\
GCA NLM                  & 0.9000                                      & 0.8467                            & 0.6752                           \\
LR-MVL                   & 0.9171                                   & 0.8683                            & 0.7323                           \\
Skip-Gram NLM        & 0.9095                                   & 0.8647                            & 0.7194                           \\
SCODE            & 0.9207                                   & \textbf{0.8835}                   & \textbf{0.7739}     
\end{tabular}
\end{center}
\end{table}

\subsection*{Dependency Parsing} \label{sec:depparsing1}

We chose CoNLL-2008 data \cite{conll2008} as the benchmark to compare word embeddings in English Dependency Parsing. For computational reasons, we fixed  the training set to the first 5K sentences of CONLL 2008 English dataset. However, we conducted experiments using full training set with SCODE embeddings. For multilingual experiments, we chose CoNLL-2006 Shared Task languages Bulgarian, Spanish, Czech, German, Swedish, and Turkish \cite{conll2006}. 

We used a framework \cite{lei2014} that is capable of incorporating word embeddings in dependency parsing. It reduces the dimensionality of head-modifier feature vectors by learning a tensor of low rank. The model is able to combine features from state-of-the-art parsers MST Parser \cite{mcdonald2005online} and Turbo Parser \cite{martins2013turning} as well as low-rank tensor features which includes word embeddings. Features used in the model is listed in Table~\ref{features-depparse}.

\begin{table}[h]
\small
\caption{\label{features-depparse} Features Used In Low-Rank Tensor based Dependency Parser }
\begin{center}
\begin{itemize}
  \item Unigram Features:  for current word $x_{i}$ form,lemma and POS tag of $x_{i,i-1,i+2}$, morphology of $x_{i}$, bias
  \item Bigram Features : previous and current POS tag, the current and next POS tag, current POS and lemma, current lemma and morphology
 \item Trigram Features: POS tag of the previous, current, and next word.
  \item Embedding features: $e_{i}[d]$ for $i$ in  \{-1,0,+1\},  where $d$ ranges over the dimensions of the embedding $e_{i}$. 
\end{itemize}
\end{center}
\end{table} 

There are two hyperparameters $\gamma$ and $r$. The first one balances tensor features and traditional MST/Turbo features. The second one is the rank of the tensor. We set the hyperparameters $\gamma = 0.3$ and $r = 50$ and ran third-order model to get comparable result in that work.

Table ~\ref{depparse-table1} shows the Unlabeled Accuracy Scores for word embeddings and the baseline parser which is not using word embeddings. Each word embedding shows improvements over baseline parser. However, improvements are marginal, similar to Chunking results. SCODE embeddings trained on RCV1 corpus achieve the best scores among others. 


\begin{table}[h]
\small
\caption{\label{depparse-table1} Dependency Parsing Results for ConLL 2008 English Data for Word Embeddings. The ones in bold font are the highest scores in their columns. }
\begin{center}
\begin{tabular}{lll}
\textbf{Word Embeddings} & \textbf{Training Score} & \textbf{Test Score} \\ \hline
Baseline & 0.9447 & 0.8976 \\
C\&W & 0.9332 & 0.9007 \\
HLBL & \textbf{0.9459} & 0.9013 \\
GCA NLM & 0.9140 & 0.8985 \\
LR-MVL & 0.9308 & 0.9016 \\
Skip-Gram NLM & 0.9397 & 0.9014 \\
SCODE & 0.9444 & \textbf{0.9028} \\

\end{tabular}
\end{center}
\end{table}

We report Multilingual Dependency Parsing scores in Table ~\ref{depparse-table2}. In the first column, the results reported in \cite{lei2014} is listed. In the second column, the state-of-the-art results before \cite{lei2014}. In the third column, the parser using the SCODE embeddings are listed. SCODE embeddings improve parsers for 6 out of 7 languages and achieve the best results for 5 out of 7 of them.

\begin{table}[h]
\small
\caption{\label{depparse-table2} Dependency Parsing Results for ConLL 2006 Languages for SCODE Embeddings. English results are from ConLL 2008. The ones in bold font are the highest scores in their rows. }
\begin{center}
\begin{tabular}{lrrr}
\textbf{Language} & \textbf{Baseline} & \textbf{State-of-The-Art} & \textbf{SCODE Embeddings} \\ \hline
Bulgarian         & 0.9350 			  & 0.9402                    & \textbf{0.9413}                    \\
Czech             & \textbf{0.9050}   & 0.9032                    & 0.9038                    \\
English           & 0.9302            & 0.9322                    & \textbf{0.9344}           \\
German            & 0.9197            & \textbf{0.9241}           & 0.9233                    \\
Spanish           & 0.8800            & 0.8796                    & \textbf{0.8823}           \\
Swedish           & 0.9100            & 0.9162                    & \textbf{0.9165}           \\
Turkish           & 0.7684            & 0.7755                    & \textbf{0.7783}          
\end{tabular}
\end{center}
\end{table}

\section{Conclusion}
\label{sec:conclusion}

We analyzed SCODE word embeddings in supervised
NLP tasks. SCODE word embeddings are previously used in unsupervised
part of speech tagging \cite{yatbaz2012,cirik2013addressing,yatbaz2014}
and word sense induction \cite{baskaya2013ai}. Their first use in a supervised setting was in dependency parsing \cite{cirik2013ai}, however, results were inconclusive. \cite{lei2014} successfully make use of SCODE embeddings as
additional features in dependency parsing.

We compared SCODE word embeddings with existing word 
embeddings in Chunking, NER, and Dependency Parsing. For all these benchmarks, we used publicly available implementations. They all are near state-of-the-art solutions in these tasks. SCODE word embeddings are at least good as other word embeddings or achieved better results. 

We analyzed SCODE embeddings in multilingual Dependency Parsing. SCODE embeddings are consistent in improving the baseline systems. Note that
other word embeddings are not studied in multilingual settings yet. SCODE word embeddings  and the code used in generating embeddings in this work is publicly available\footnote{link}.

\bibliographystyle{natbib}
\bibliography{nips2014}

\end{document}